\documentclass[conference]{IEEEtran}
\IEEEoverridecommandlockouts
\usepackage{cite}
\usepackage{amsmath,amssymb,amsfonts}
\usepackage{algorithmic}
\usepackage{subcaption}
\usepackage[export]{adjustbox}
\usepackage{rotating} % 用于旋转图片
\usepackage{graphicx}
\usepackage{textcomp}
\usepackage{multirow}
\usepackage{xcolor}
\usepackage{booktabs} % 用于绘制更专业的表格
\usepackage{amsmath} % 用于数学公式
\usepackage{makecell} % 用于表格单元格内换行
\definecolor{bestcolor}{RGB}{255, 50, 50} % 荧光红色
\definecolor{secondbest}{RGB}{0, 150, 255} % 荧光蓝色
\definecolor{thirdbest}{RGB}{0, 200, 0} % 荧光绿色
\def\BibTeX{{\rm B\kern-.05em{\sc i\kern-.025em b}\kern-.08em
    T\kern-.1667em\lower.7ex\hbox{E}\kern-.125emX}}
\usepackage[colorlinks=True, citecolor=blue, linkcolor=blue, urlcolor=black]{hyperref}
\begin{document}

\title{SEP-YOLO: Fourier-Domain Feature Representation for Transparent Object Instance Segmentation\\
}

\author{\IEEEauthorblockN{Fengming Zhang}
\IEEEauthorblockA{
\textit{Jiangnan University}\\
Wuxi, China \\
6233110051@stu.jiangnan.edu.cn}
\and
\IEEEauthorblockN{Tao Yan*}
\IEEEauthorblockA{
\textit{Jiangnan University}\\
Wuxi, China \\
yantao@jiangnan.edu.cn}
\and
\IEEEauthorblockN{Jianchao Huang}
\IEEEauthorblockA{
\textit{Jiangnan University}\\
Wuxi, China \\
6233112018@stu.jiangnan.edu.cn}
\thanks{Tao Yan* is Corresponding author.}
}

\maketitle

\begin{abstract}
Transparent object instance segmentation presents significant challenges in computer vision, due to the inherent properties of transparent objects, including boundary blur, low contrast, and high dependence on background context. Existing methods often fail as they depend on strong appearance cues and clear boundaries. To address these limitations, we propose SEP-YOLO, a novel framework that integrates a dual-domain collaborative mechanism for transparent object instance segmentation. Our method incorporates a Frequency Domain Detail Enhancement Module, which separates and enhances weak high-frequency boundary components via learnable complex weights. We further design a multi-scale spatial refinement stream, which consists of a Content-Aware Alignment Neck and a Multi-scale Gated Refinement Block, to ensure precise feature alignment and boundary localization in deep semantic features. We also provide high-quality instance-level annotations for the Trans10K dataset, filling the critical data gap in transparent object instance segmentation. Extensive experiments on the Trans10K and GVD datasets show that SEP-YOLO achieves state-of-the-art (SOTA) performance.
\end{abstract}


\section{Introduction}
Transparent objects, such as architectural glass windows, laboratory glassware, and drinking glasses, are ubiquitous in daily life. Accurate detection and segmentation of these objects are critical for diverse applications, such as robotic manipulation, autonomous driving, and industrial defect inspection\cite{r1}. While significant progress has been made in semantic segmentation of transparent objects\cite{r2,r3,r4}, including Transformer-based Trans4Trans\cite{r5} and boundary-enhanced EBLNet\cite{r7}, these methods inherently cannot distinguish between different instances of the same category\cite{r22}, thus limiting their applicability in robotic grasping and industrial sorting\cite{r25}. In contrast, instance segmentation addresses this limitation by locating transparent object regions and delineating precise boundaries for each distinct instance\cite{r23}.

Although numerous instance segmentation methods have been proposed\cite{r8,r9,r11,r16}, few are designed specifically for transparent objects. Transparent objects exhibit unique physical properties, including high light transmittance and low reflectivity. As a result, their appearance is strongly dependent on the background, with no distinctive texture or color features. Furthermore, complex light refraction results in severely blurred boundaries that merge with the background\cite{r12,r13,r14}, and these challenges lead to significant performance degradation in conventional instance segmentation models, which rely on strong appearance cues and clear boundaries.

Recently, Cherian et al.\cite{r15} proposed TrInSeg, a method that achieves data-efficient transparent instance segmentation under a few-shot setting using TransMixup data augmentation and template consistency filtering, and shows  improved performance in robotic bin-picking tasks. However, this method  relies on the assumption that transparent objects are rigid and regularly shaped, which considerably limits its generalization capability to non-rigid or irregular transparent objects. 
\begin{figure}[htbp]
    \centering
    \includegraphics[width=0.5\textwidth, max height=8cm]{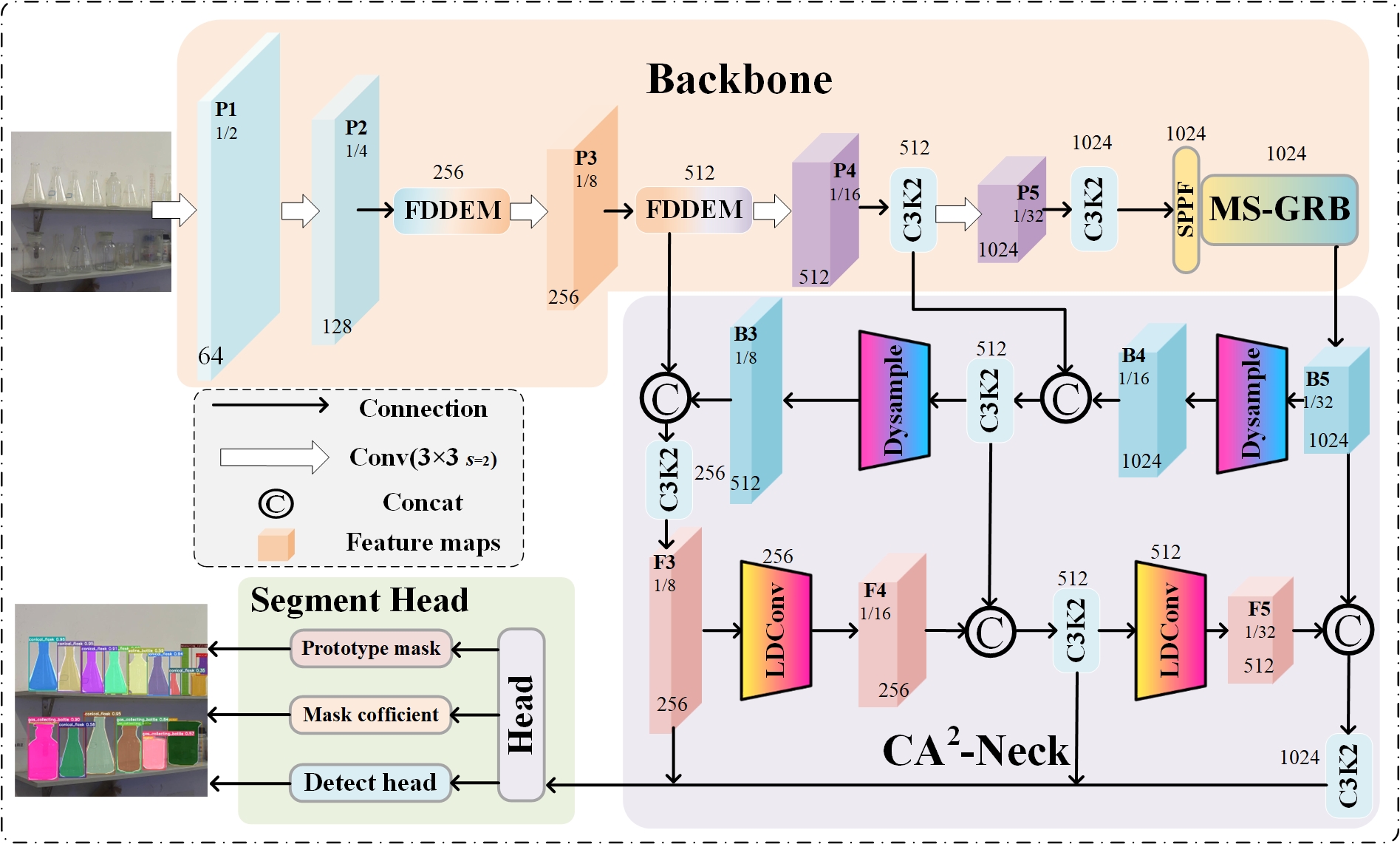}
    \caption{SEP-YOLO framework based on YOLO11}
    \label{sep}
\end{figure}

To address the above issues, we propose the SEP-YOLO framework as shown in Fig.~\ref{sep}, with the following main contributions:

\begin{itemize}
    \item We propose the Frequency Domain Detail Enhancement Module (FDDEM), which enhances weak boundary components of transparent objects via learnable complex weights in the frequency domain, addressing the low signal-to-noise ratio issue in spatial domain features.

    \item We design a multi-scale spatial refinement stream, consisting of $\text{CA}^{2}\text{-Neck}$ and $\text{MS-GRB}$. This stream enables precise cross-scale feature fusion through content-aware alignment and dynamic gating, facilitating accurate boundary localization and noise suppression.

    \item We provide high-quality instance-level annotations for the $\text{Trans10K}$ dataset\cite{r21}, filling the instance segmentation data gap for daily scenes. Our method achieves SOTA performance on both $\text{Trans10K}$ and $\text{GVD}$ while maintaining real-time inference speed, demonstrating strong potential for industrial applications.
\end{itemize}
\section{Proposed Method}

\subsection{Frequency Domain Detail Enhancement Module}

The inherent physical properties of transparent objects result in severely blurred boundaries, which manifest as high-frequency components with extremely low signal-to-noise ratio. These fragile details are easily diluted during convolution and pooling, leading to irreversible boundary information loss. To address this, we propose the Frequency Domain Detail Enhancement Module (FDDEM), which enhances frequency components corresponding to transparent object boundaries through learnable frequency domain complex weights. The refined components are then reintegrated into the spatial domain, providing clearer and more discriminative feature representations. FDDEM employs a dual-branch structure. The spatial context branch extracts and preserves multi-scale contextual information through convolutional layers,  while the frequency detail branch first maps input features into the frequency domain using Fast Fourier Transform (FFT)\cite{r24}. The transformation is formulated as:

\begin{equation}
\mathcal{F}(\mathbf{X})(u,v) = \sum_{x=0}^{H-1}\sum_{y=0}^{W-1} \mathbf{X}(x,y) \cdot e^{-j2\pi\left(\frac{ux}{H} + \frac{vy}{W}\right)},
\end{equation}
where $\mathbf{X} \in \mathbb{R}^{H \times W \times C}$ denotes the input feature map, and $\mathcal{F}(\mathbf{X})$ represents its frequency domain representation.

\begin{figure}[htbp]
    \centering
    \subfloat[FDDEM]{\includegraphics[width=0.48\textwidth, max height=6.5cm]{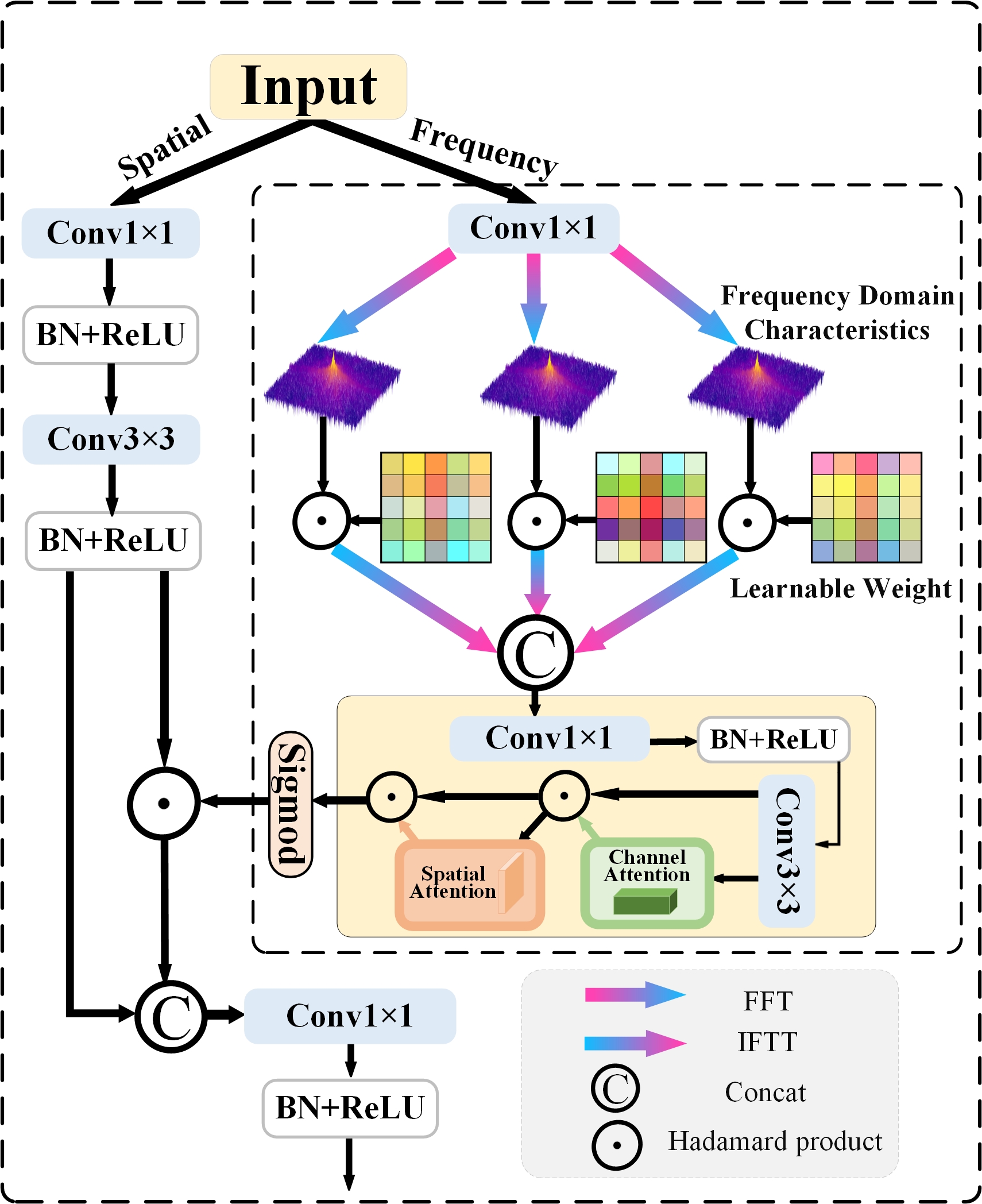}}
    \hfill
    \subfloat[MS-GRB]{\includegraphics[width=0.48\textwidth, max height=6.5cm]{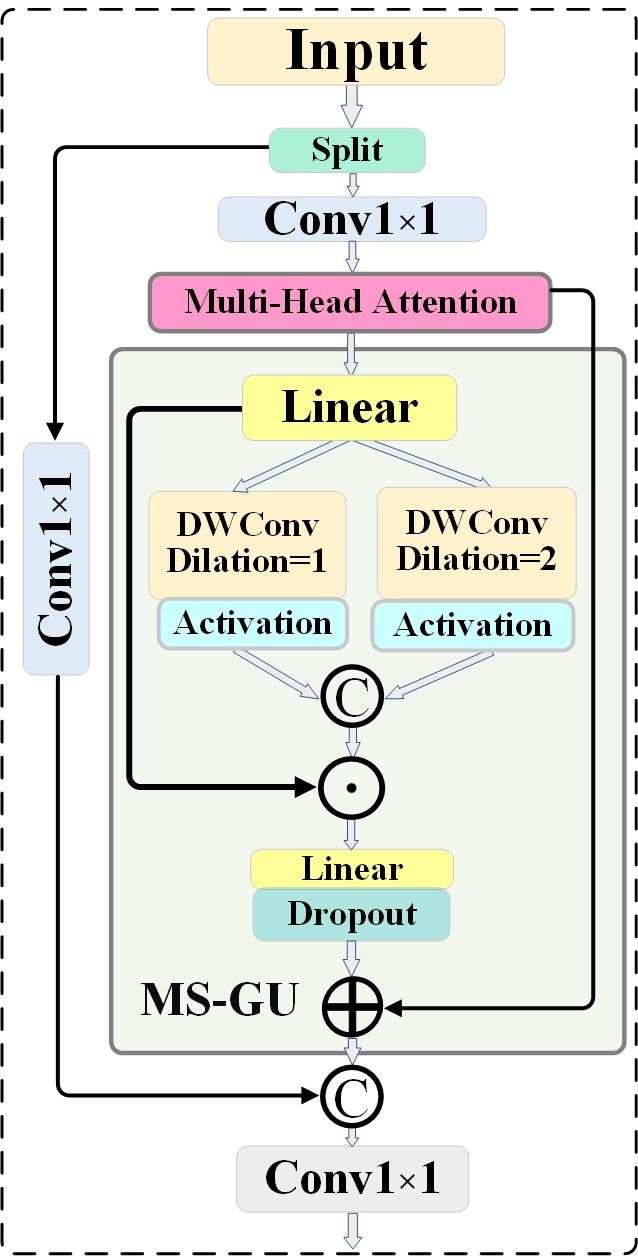}}
    \caption{Architecture of the proposed modules: (a) FDDEM; (b) MS-GRB}
    \label{F-G}
\end{figure}

We employ a multi-branch frequency enhancement strategy that adaptively modulates frequency domain features through learnable complex weight matrices. Unlike conventional high-pass filters with fixed frequency responses, our approach enables the network to autonomously discover optimal enhancement patterns tailored to the specific characteristics of transparent object boundaries. Each branch utilizes distinct complex weight matrices to adjust spectral components via element-wise multiplication, as illustrated in Fig.~\ref{F-G}(a). The frequency enhancement process for each branch is defined as:

\begin{equation}
\mathcal{F}_{\text{enhanced}}^i = \mathcal{F}(\mathbf{X}) \odot \mathbf{W}^i,
\end{equation}
where $\mathbf{W}^i \in \mathbb{C}^{C \times H \times W}$  are learnable complex weight matrices, and $\odot$ denotes the  Hadamard product. The real components of $\mathbf{W}^i$ control amplitude modulation, while the imaginary components adjust phase relationships, providing comprehensive control over frequency domain characteristics.

These adaptively enhanced frequency components are transformed back to the spatial domain via Inverse Fast Fourier Transform (IFFT). The enhanced features from all branches are concatenated and compressed via a 1×1 convolution. A dual-attention mechanism then generates a frequency-guided attention map, which adaptively weights and enhances the spatial features. The frequency-enhanced features are then integrated with the spatial branch. This integration operation connects the boundary-refined information from the frequency domain with the contextual representations from the spatial domain, forming a unified feature representation that combines enhanced boundary details with rich spatial context. 

\subsection{Multi-Scale Gated Refinement Block}
While our FDDEM module enhances high-frequency boundaries in the frequency domain, features become diluted through spatial downsampling and aggregation, leading to a loss of boundary details and inaccurate localization in semantic features. To address this, we propose the Multi-scale Gated Refinement Module (MS-GRB) as shown in Fig.~\ref{F-G}(b).

The core component is the Multi-scale Gating Unit (MS-GU), essentially a multi-scale variant of the Convolutional Gated Linear Unit (CGLU)\cite{r26}. It employs MSDWConv to aggregate multi-scale depthwise convolutions for efficient context extraction, while ensuring computational efficiency. As a multi-scale CGLU, it enhances the non-linear representation capability of spatially extracted features and optimizes cross-scale feature refinement; the gating mechanism\cite{r17} further performs adaptive channel-wise weighting and noise suppression on these refined features.
 The complete MS-GRB operation integrates MS-GU with residual learning:
\begin{equation}
\label{eq:msgrb_final}
\mathbf{Y} = \mathbf{X} + \text{Conv}_{1 \times 1}^{\text{shrink}} \left( \mathcal{D}_{\text{MS}} \left( \sigma_{\text{act}}(\mathbf{X}_k) \right) \odot \sigma_{\text{gate}}(\mathbf{V}_k) \right),
\end{equation}
where $\mathbf{X}_k$ and $\mathbf{V}_k$ are feature branches from channel splitting, $\mathcal{D}_{\text{MS}}(\cdot)$ is MSDWConv, and $\sigma_{\text{act}}(\cdot)$/$\sigma_{\text{gate}}(\cdot)$ are activation and gating functions. This deep gated refinement enables MS-GRB to achieve precise localization and enhancement of faint boundary information at the deepest semantic level, significantly improving segmentation accuracy and generalization in complex backgrounds.

\subsection{Content-Aware Alignment Neck}
Due to the optical properties of transparent objects, boundaries appear blurred and spatial positions become unstable. In feature pyramid structures, both downsampling and upsampling operations face limitations: downsampling truncates feature information and reduces spatial details, while upsampling averages pixel values, diluting high-frequency boundary information and causing spatial misalignment. To address these challenges, we propose the Content-Aware Alignment Neck (CA\textsuperscript{2}-Neck) with dual-path enhancements.

\begin{figure}[htbp]
    \centering
    \subfloat[Illustration of sampling position generation process in LDConv with N=5]{\includegraphics[width=0.45\textwidth]{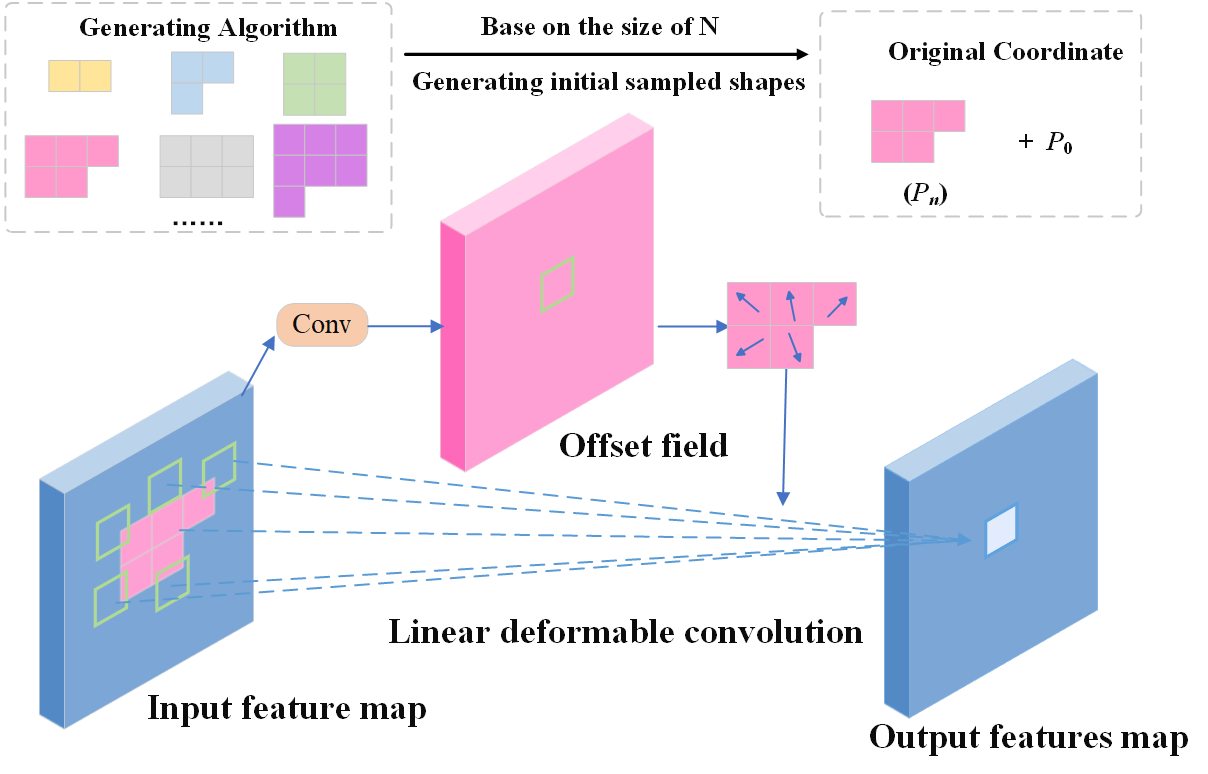}}
    \hfill
    \subfloat[DySample with static scope factor]{\includegraphics[width=0.45\textwidth]{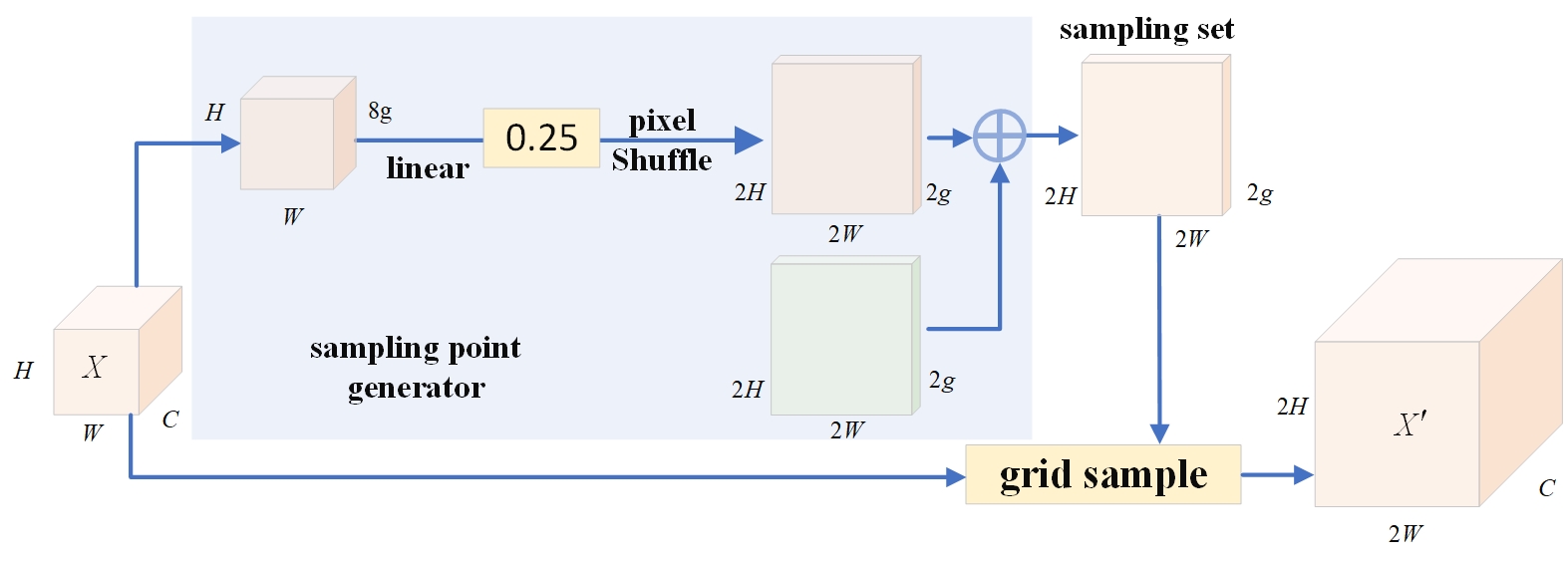}}
    \caption{(a) LDConv; (b) DySample}
    \label{L-D}
\end{figure}

In the downsampling path, we employ the Linear Deformable Convolution (LDConv) to replace standard strided convolutions. As shown in Fig.~\ref{L-D}(a), LDConv generates convolutional kernels with arbitrary numbers of parameters and arbitrary sampling shapes through a novel coordinate generation algorithm. This enables linear rather than quadratic parameter growth while maintaining representation capacity. The LDConv operation adapts to varying targets through learnable offsets that dynamically adjust sampling positions:

\begin{equation}
\mathbf{Y}_{\text{LDConv}} = \sum_{n=1}^{N} w_n \cdot x(\mathbf{P}_0 + \mathbf{P}_n + \Delta\mathbf{P}_n),
\end{equation}
where $N$ is the number of convolutional parameters, $\mathbf{P}_n$ are the initial sampling coordinates generated by the algorithm, and $\Delta\mathbf{P}_n$ are the learned offsets. This formulation enables LDConv to capture extensive global context while preserving spatial details during downsampling, with parameter count growing linearly with kernel size rather than quadratically as in standard and deformable convolutions.

In the upsampling path, we adopt DySample \cite{r18}, an innovative dynamic upsampler that reformulates the upsampling process from the perspective of point sampling as shown in Fig.~\ref{L-D}(b). The core operation can be expressed as:
\begin{equation}
\mathbf{X}' = \text{grid\_sample}\left(\mathbf{X}, \mathbf{G} + \lambda \cdot \text{linear}(\mathbf{X})\right),
\end{equation}
where $\mathbf{G}$ is the original sampling grid and $\lambda=0.25$ is the static scope factor. This design employs bilinear initialization to ensure uniform distribution of initial sampling positions, while the scope factor constrains the offset range to prevent overlap between adjacent sampling points, effectively avoiding boundary artifacts and spatial misalignment. This mechanism enables DySample to adaptively adjust sampling positions based on feature content, generating semantically responsive sampling points in texture-rich edge regions to better preserve detailed information.

The dual-path enhancement collectively maintains boundary details and ensures accurate multi-scale feature alignment throughout the pyramid structure, particularly beneficial for transparent objects with blurred boundaries and unstable spatial positions.

\section{Experiments}

\subsection{Datasets}\label{AA}
 We evaluate SEP-YOLO on two representative transparent object datasets: GVD\cite{r29} and Trans10k\cite{r21}. The GVD dataset comprises 2,416 laboratory scenes, covering 14 classes of transparent chemical instruments. Since existing instance segmentation datasets lack coverage of common transparent household items, we utilized the Trans10K semantic segmentation dataset, adding instance-level annotations to adapt it for the instance segmentation task. This adapted dataset includes 9,491 images, encompassing two transparent object categories: glass surfaces and glassware. These two datasets collectively cover most scenarios encountered in daily life and professional laboratories. Comparison of SEP-YOLO with state-of-the-art detection methods demonstrates that SEP-YOLO achieves superior performance in transparent object segmentation.
     \begin{table*}[t]
      \centering
      \caption{Performance comparison of different methods on Trans10k and GVD datasets. The best, second best, and third best results are highlighted in \textcolor{bestcolor}{red}, \textcolor{secondbest}{blue}, and \textcolor{thirdbest}{green}, respectively.}
      \label{tab:results}
      \begin{tabular}{lccccccccccc}
        \toprule
        \multirow{2}{*}{\textbf{Method}} & \multirow{2}{*}{\textbf{Venue}} & \multicolumn{2}{c}{Box $mAP_{50}$ } & \multicolumn{2}{c}{Box $mAP_{75}$ } & \multicolumn{2}{c}{Mask $mAP_{50}$ } & \multicolumn{2}{c}{Mask $mAP_{75}$ } & \multirow{2}{*}{\begin{tabular}[c]{@{}c@{}}Params\\(M)\end{tabular}} & \multirow{2}{*}{FPS} \\
        \cmidrule(lr){3-4} \cmidrule(lr){5-6} \cmidrule(lr){7-8} \cmidrule(lr){9-10}
         & & Trans10k & GVD & Trans10k & GVD & Trans10k & GVD & Trans10k & GVD & & \\
        \midrule
        Mask-RCNN\cite{r8} & ICCV 2017 & 0.773 & 0.797 & 0.691 & 0.704 & 0.771 & 0.791 & 0.687 & 0.679 & 44.5 & 27 \\
        
        Solov2\cite{r10} & NeurIPS 2020 & 0.803 & \textcolor{secondbest}{0.856} & 0.601 & 0.642 & 0.805 & \textcolor{secondbest}{0.852} & 0.600 & 0.621 & 36.1 & 43 \\
        YOLOv10n\cite{r31} & NeurIPS 2024 & 0.794 & 0.812 & 0.689 & 0.725 & 0.798 & 0.804 & 0.691 & 0.701 & \textcolor{bestcolor}{2.64} & 87 \\
        YOLO11n & - & \textcolor{secondbest}{0.816} & 0.851 & \textcolor{secondbest}{0.742} & 0.751 & \textcolor{secondbest}{0.813} & 0.849 & \textcolor{thirdbest}{0.729} & \textcolor{secondbest}{0.742} & \textcolor{thirdbest}{2.87} & \textcolor{bestcolor}{97} \\
        TrInSeg\cite{r15} & IROS 2024 & \textcolor{thirdbest}{0.810} & 0.851 & 0.739 & 0.753 & 0.805 & 0.847 & \textcolor{secondbest}{0.737} & 0.739 & 55.6 & 21 \\
        Hyper-YOLO-N\cite{r20} & TPAMI 2024 & 0.801 & 0.819 & 0.731 & 0.740 & 0.799 & 0.809 & 0.720 & 0.732 & 3.8 & 78 \\
        Mamba-YOLO-T\cite{r19} & AAAI 2025 & 0.791 & \textcolor{thirdbest}{0.854} & 0.727 & \textcolor{secondbest}{0.760} & 0.787 & \textcolor{thirdbest}{0.850} & 0.719 & 0.733 & 6.3 & 69 \\

         YOLOv12n\cite{r30}  & arxiv 2025 & \textcolor{thirdbest}{0.810} & 0.847 & \textcolor{thirdbest}{0.741} & \textcolor{thirdbest}{0.755} & \textcolor{thirdbest}{0.807} & 0.801 & 0.711 & \textcolor{thirdbest}{0.741} & \textcolor{secondbest}{2.85} & \textcolor{secondbest}{94} \\
        \midrule
        \textbf{Ours} & \ & \textbf{\textcolor{bestcolor}{0.852}} & \textbf{\textcolor{bestcolor}{0.882}} & \textbf{\textcolor{bestcolor}{0.774}} & \textbf{\textcolor{bestcolor}{0.796}} & \textbf{\textcolor{bestcolor}{0.851}} & \textbf{\textcolor{bestcolor}{0.872}} & \textbf{\textcolor{bestcolor}{0.754}} & \textbf{\textcolor{bestcolor}{0.769}} & \textbf{2.98} & \textbf{\textcolor{thirdbest}{88}} \\
        \bottomrule
        \label{I}
      \end{tabular}
    \end{table*}
\subsection{Evaluation Metrics and Implementation Details}
Evaluation metrics include Precision, Recall, and mean Average Precision (mAP) for both bounding boxes (box) and segmentation masks (mask). SEP-YOLO was implemented using PyTorch 2.7.1, with an input image size of 640×640 and a batch size of 4, trained for 300 epochs. The Stochastic Gradient Descent (SGD) optimizer was used, with an initial learning rate of 0.0001, employing a cosine learning rate scheduler and a 3-epoch warmup phase. All experiments were conducted on a server equipped with a  NVIDIA RTX 4090 GPU and an Intel(R) Core(TM) i9-14900KF CPU.

\subsection{Comparison with State-of-the-Art Methods}
We compared SEP-YOLO with eight state-of-the-art methods, including YOLO11, on two benchmark datasets. As summarized in Table~\ref{I}, SEP-YOLO achieved the best performance on Trans10K, exceeding the second-best approach by 3.6\% in Box mAP50, 3.2\% in Box mAP75, 3.8\% in Mask mAP50, and 2.5\% in Mask mAP75. Consistent improvements were also observed on the GVD dataset across all metrics. These results confirm the effectiveness of SEP-YOLO in accurately identifying object boundaries and producing high-quality segmentation masks. In terms of efficiency, while YOLO11 remains the fastest and most compact, SEP-YOLO attains notably higher accuracy with only 0.23M additional parameters, thus maintaining a favorable balance between performance and complexity.

\begin{figure}[t]
    \centering
    \scriptsize
    % 用tabular固定列宽，确保每行对齐
    \begin{tabular}{c@{\hskip 0.005\textwidth}c@{\hskip 0.005\textwidth}c@{\hskip 0.005\textwidth}c@{\hskip 0.005\textwidth}c@{\hskip 0.005\textwidth}c}
        % 第一行图片
        \includegraphics[width=0.075\textwidth]{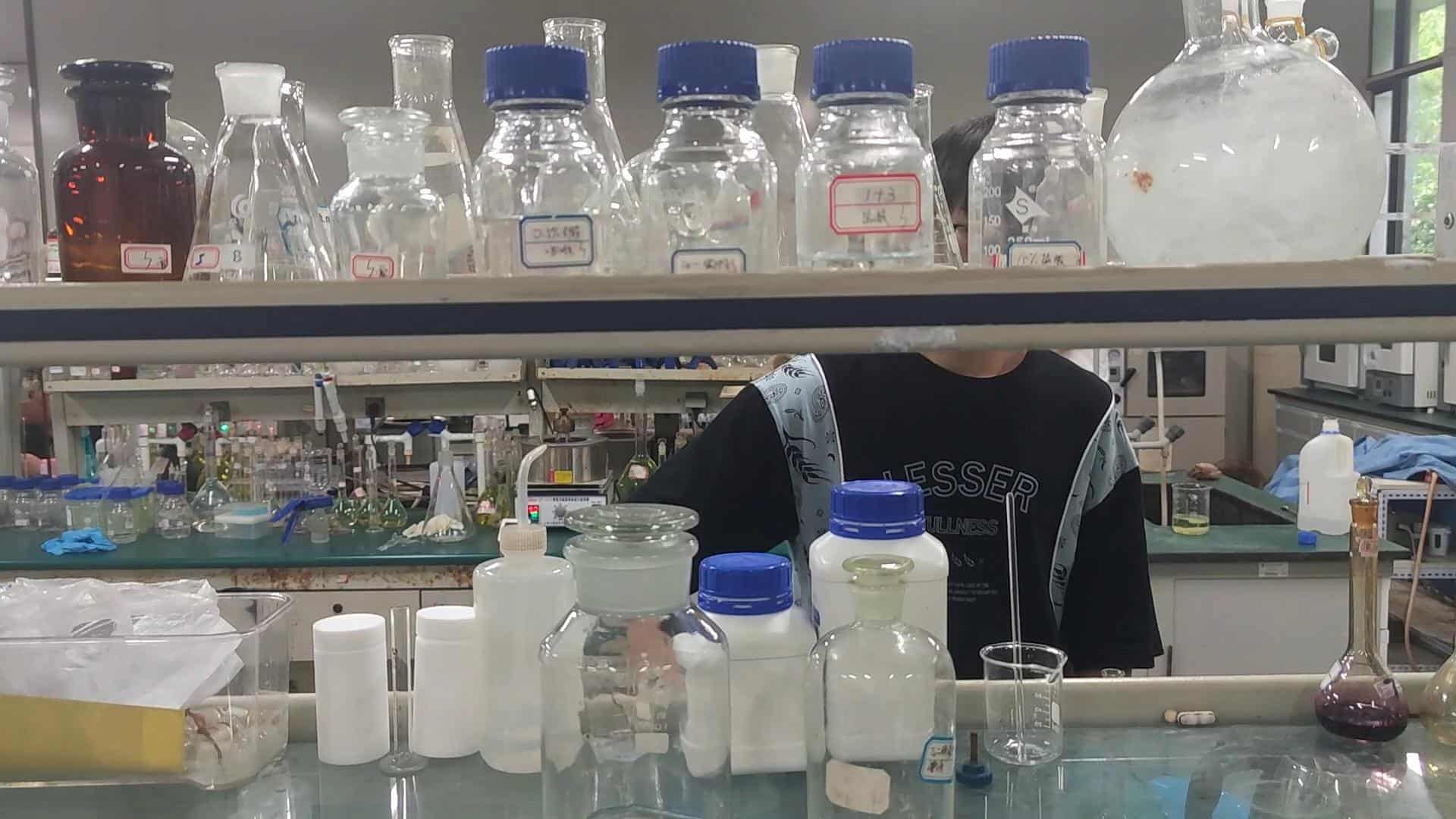} &
        \includegraphics[width=0.075\textwidth]{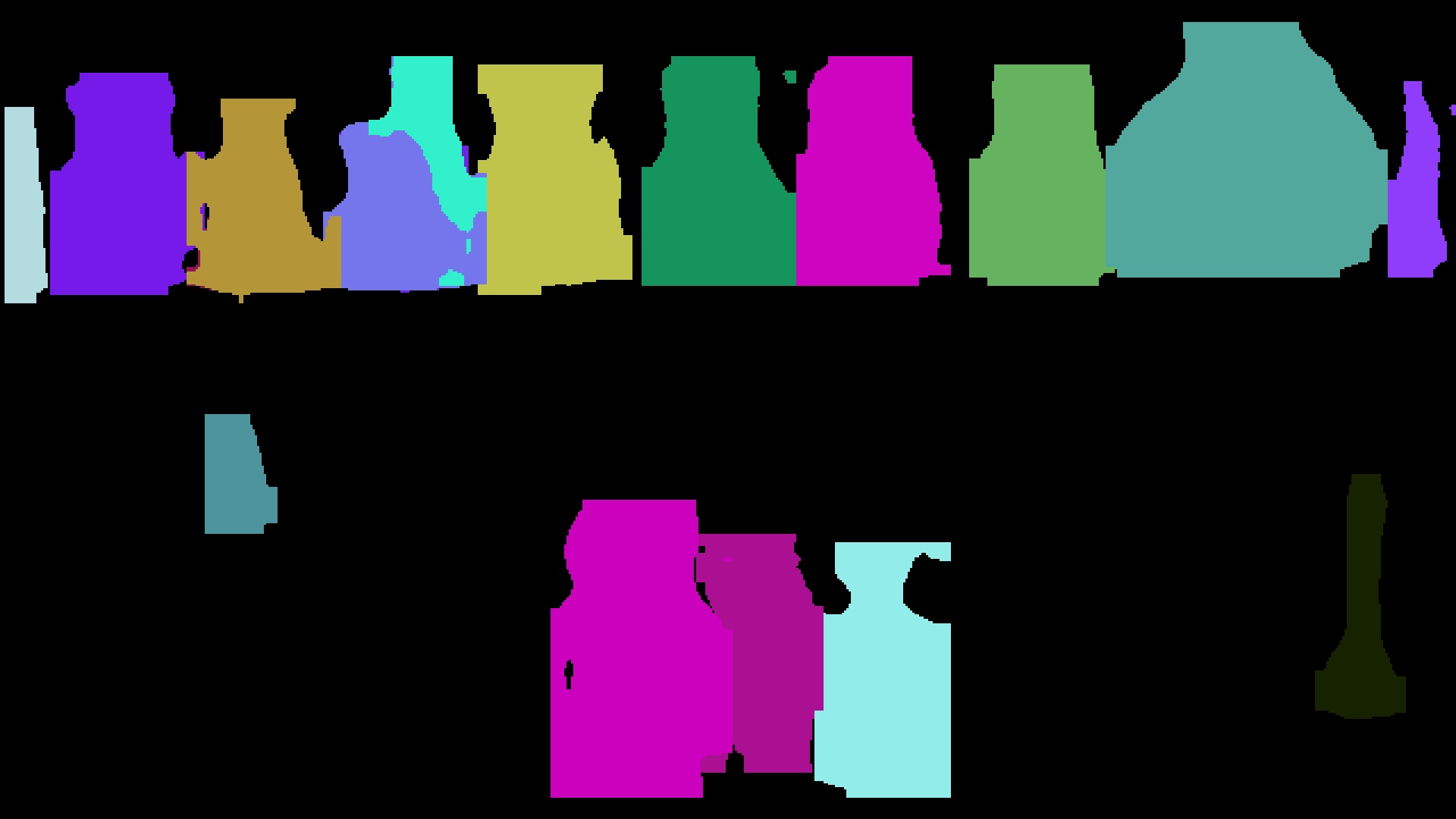} &
        \includegraphics[width=0.075\textwidth]{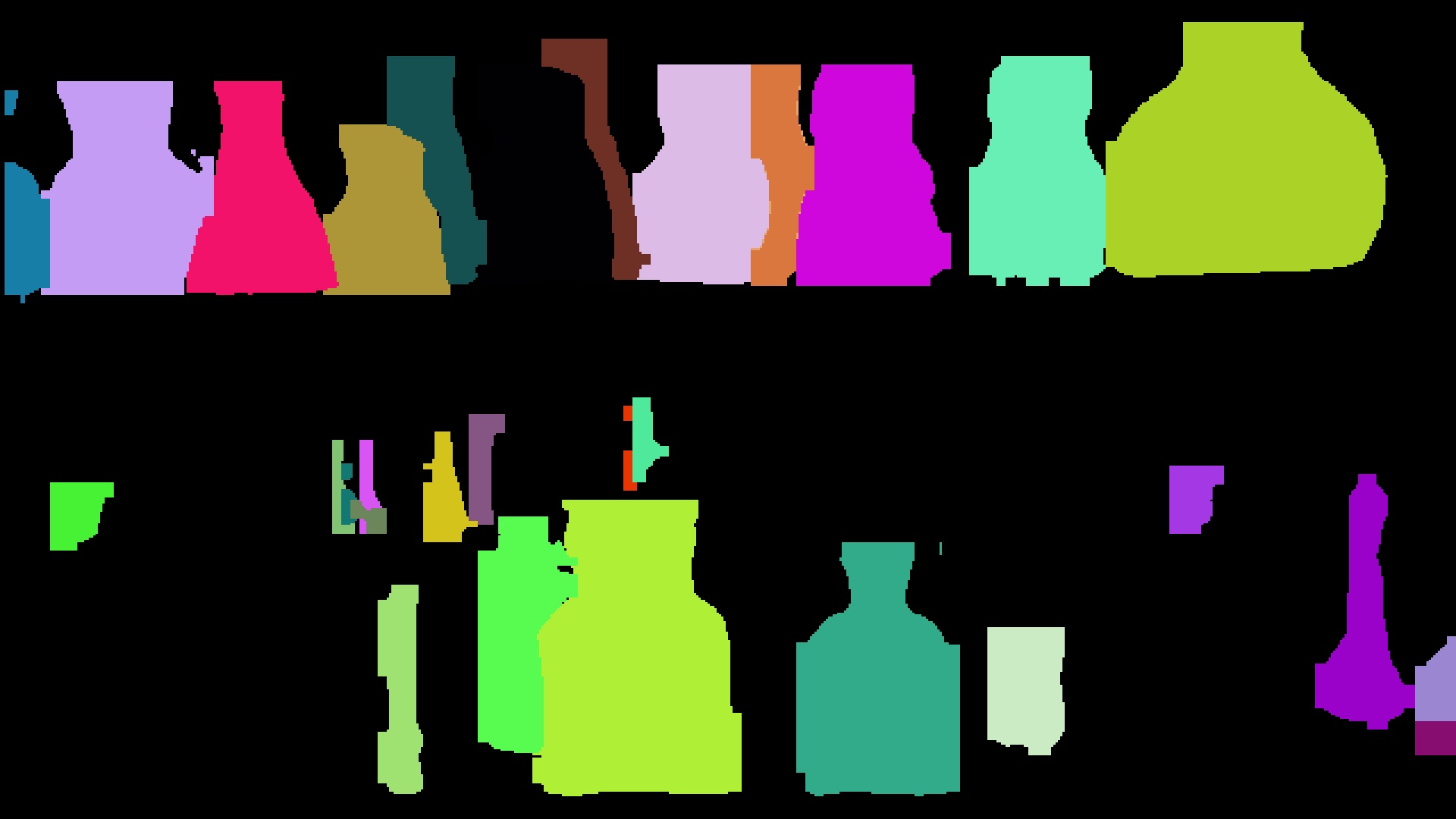} &
        \includegraphics[width=0.075\textwidth]{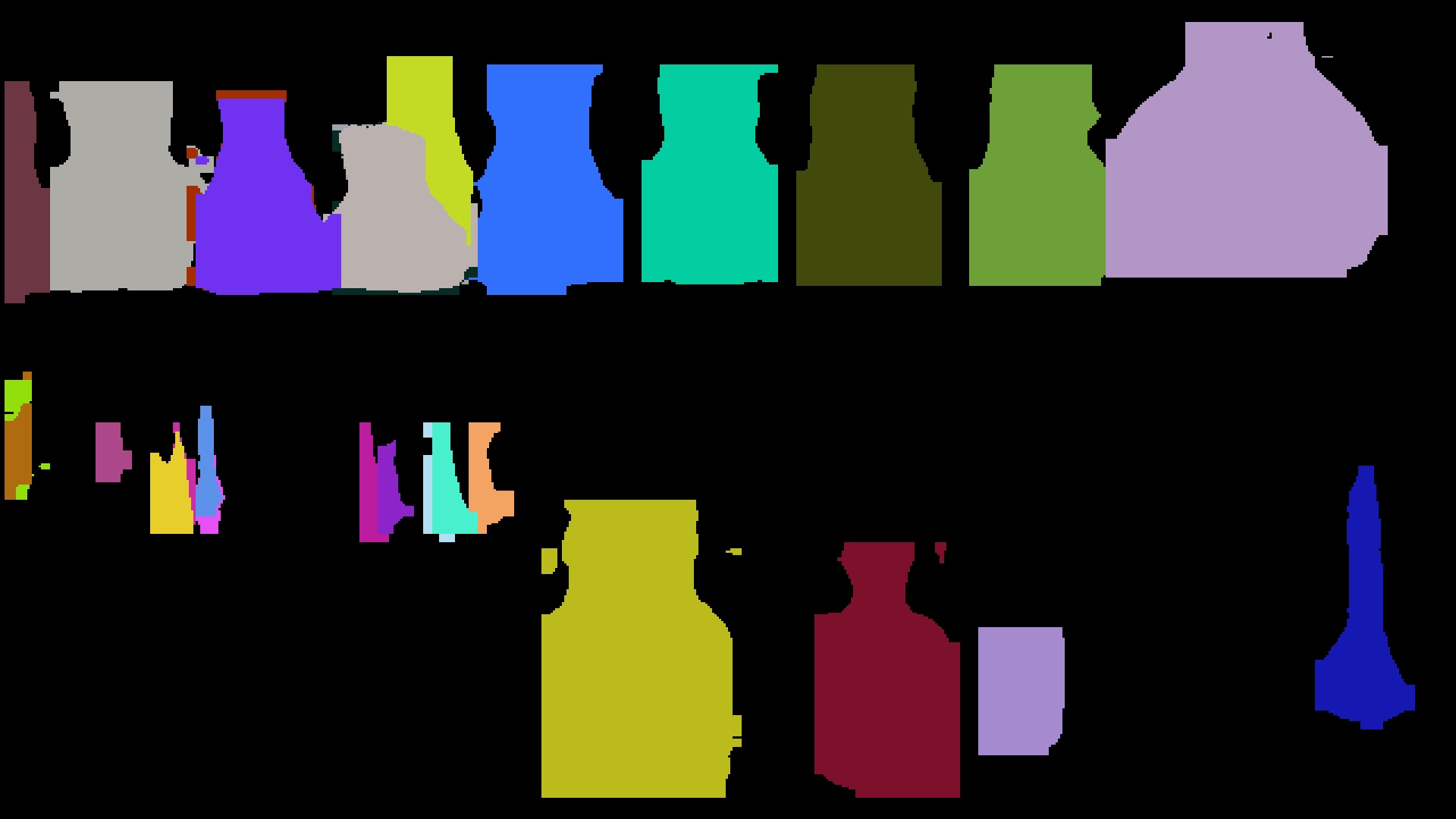} &
        \includegraphics[width=0.075\textwidth]{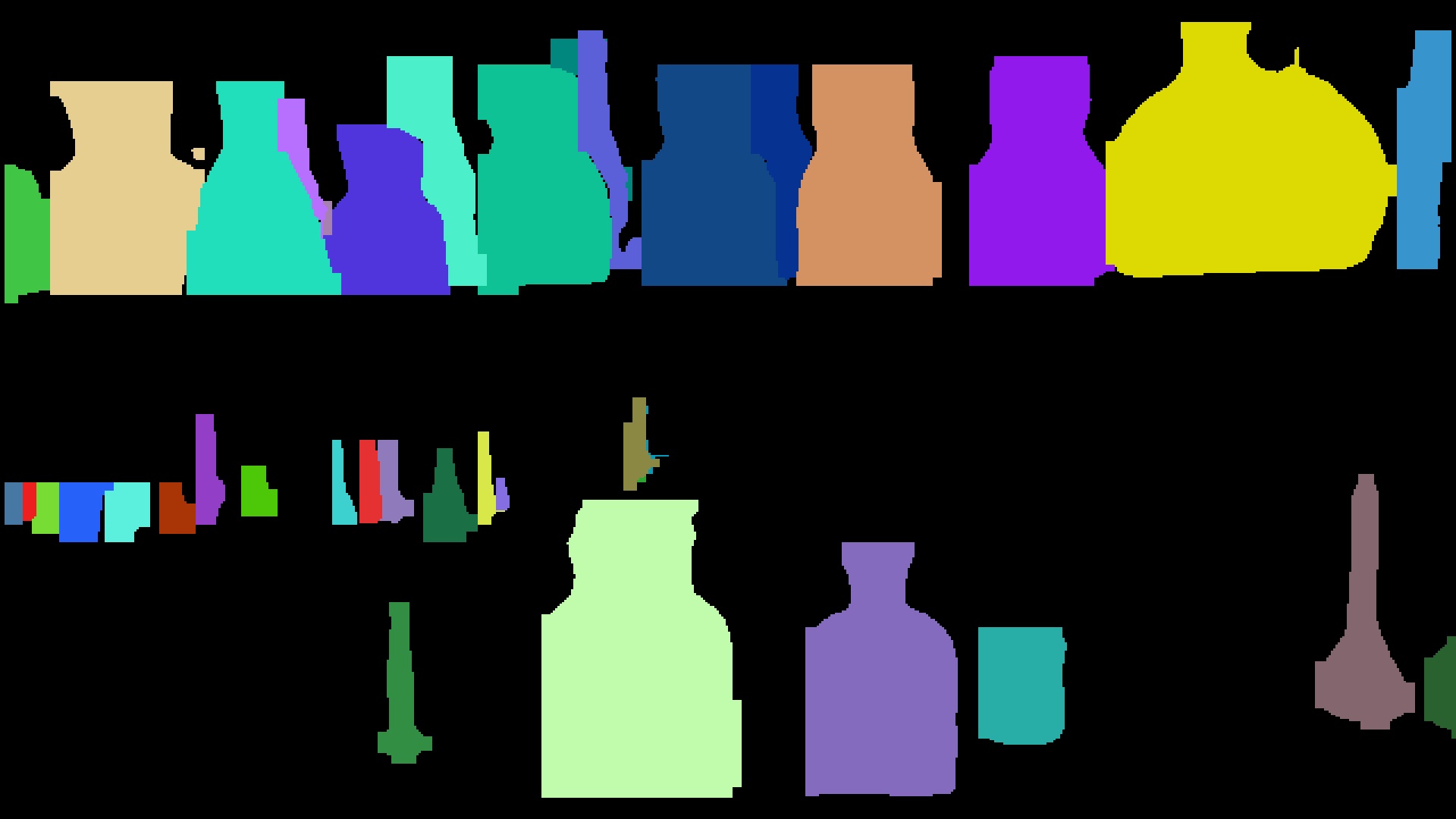} &
        \includegraphics[width=0.075\textwidth]{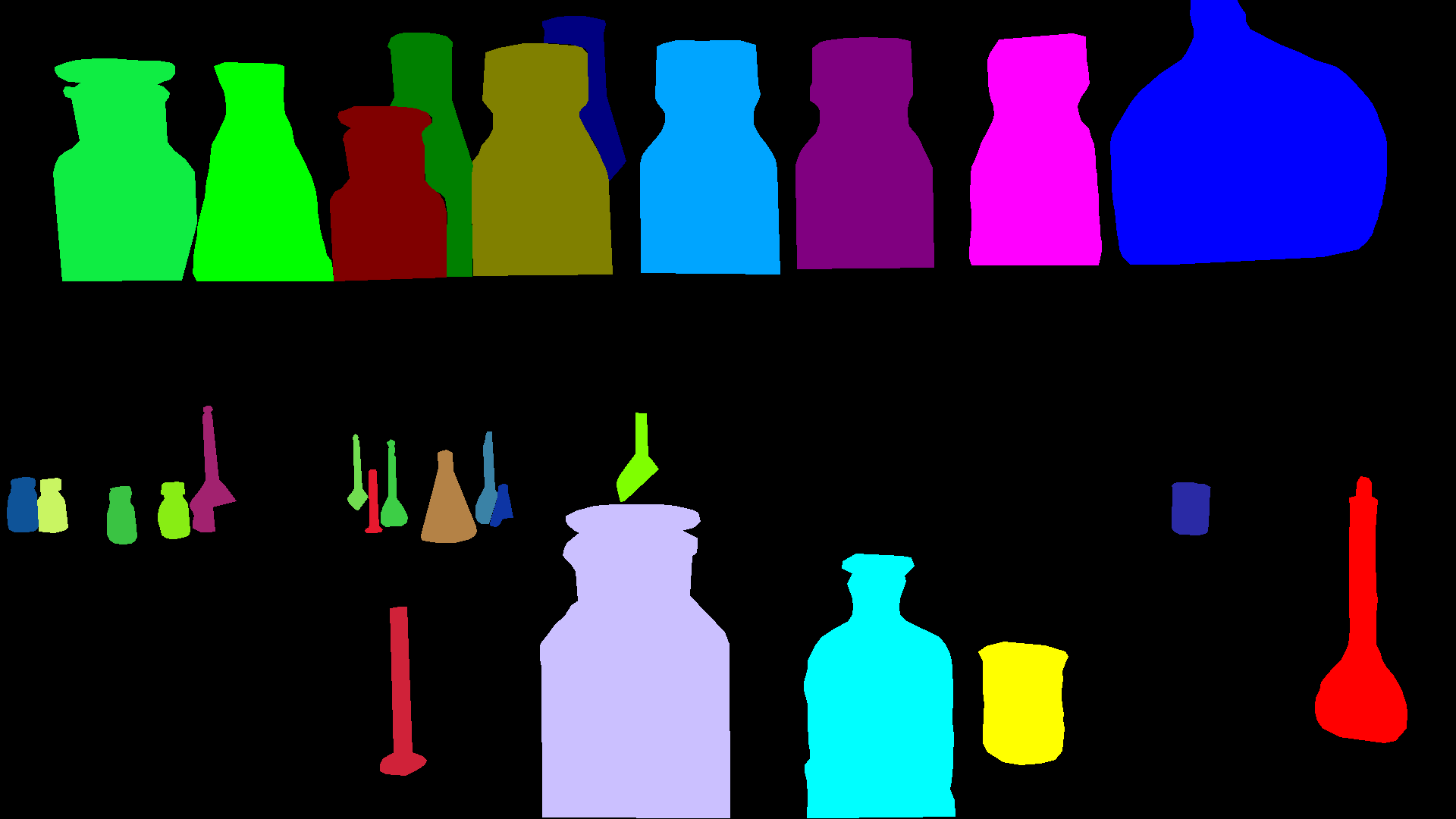} \\
        % 行间距
        
        % 第二行图片
        \includegraphics[width=0.075\textwidth]{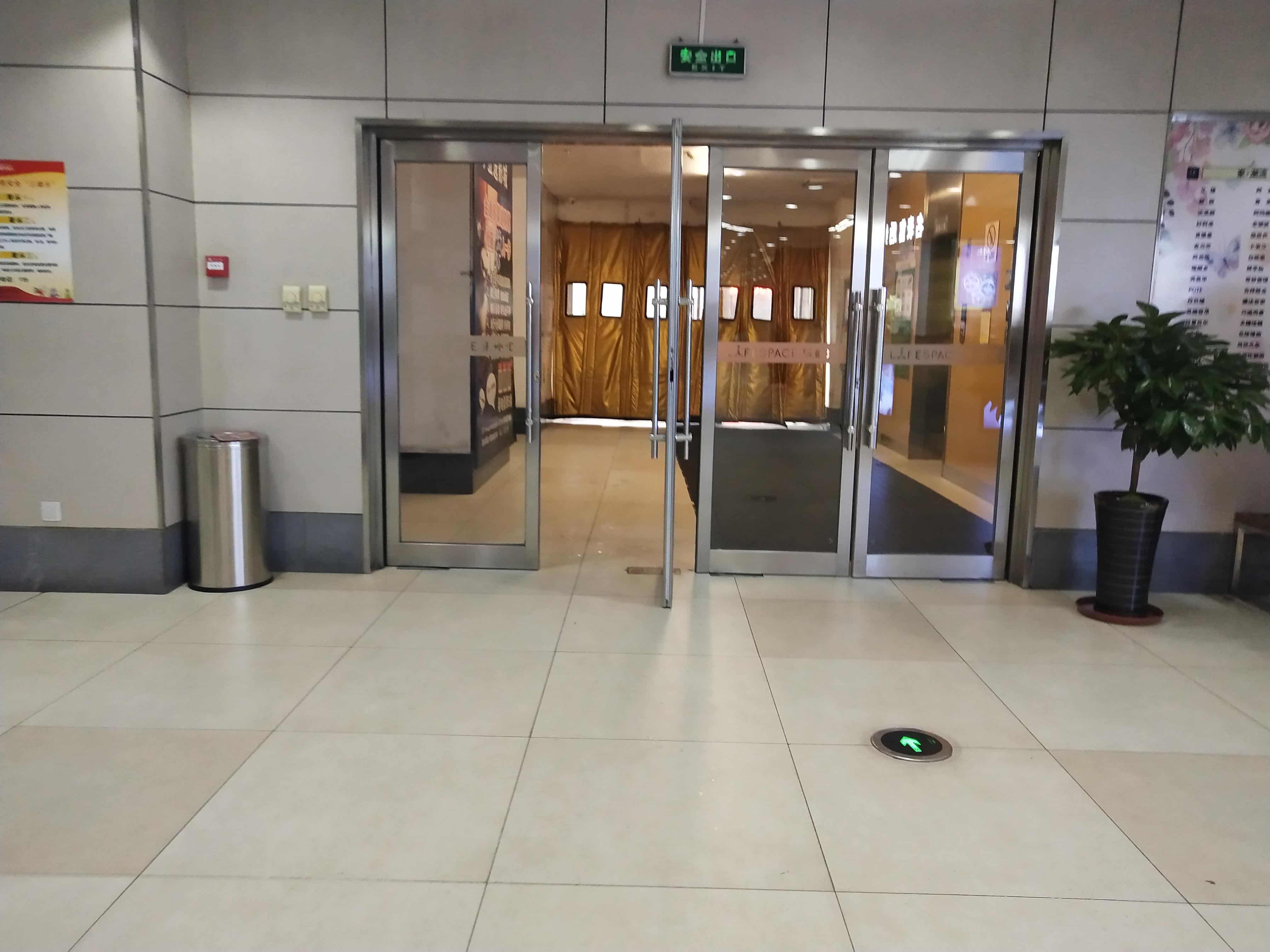} &
        \includegraphics[width=0.075\textwidth]{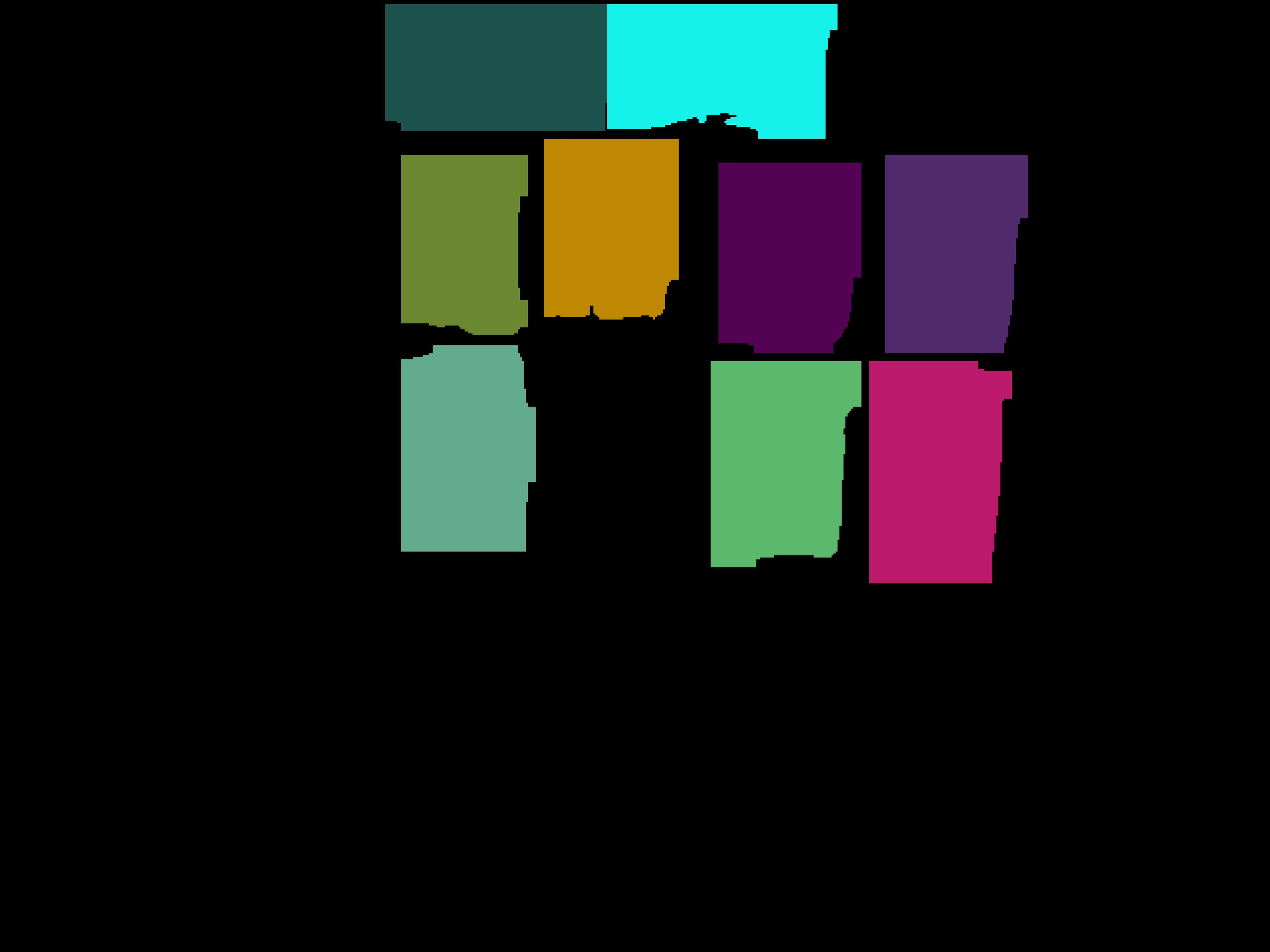} &
        \includegraphics[width=0.075\textwidth]{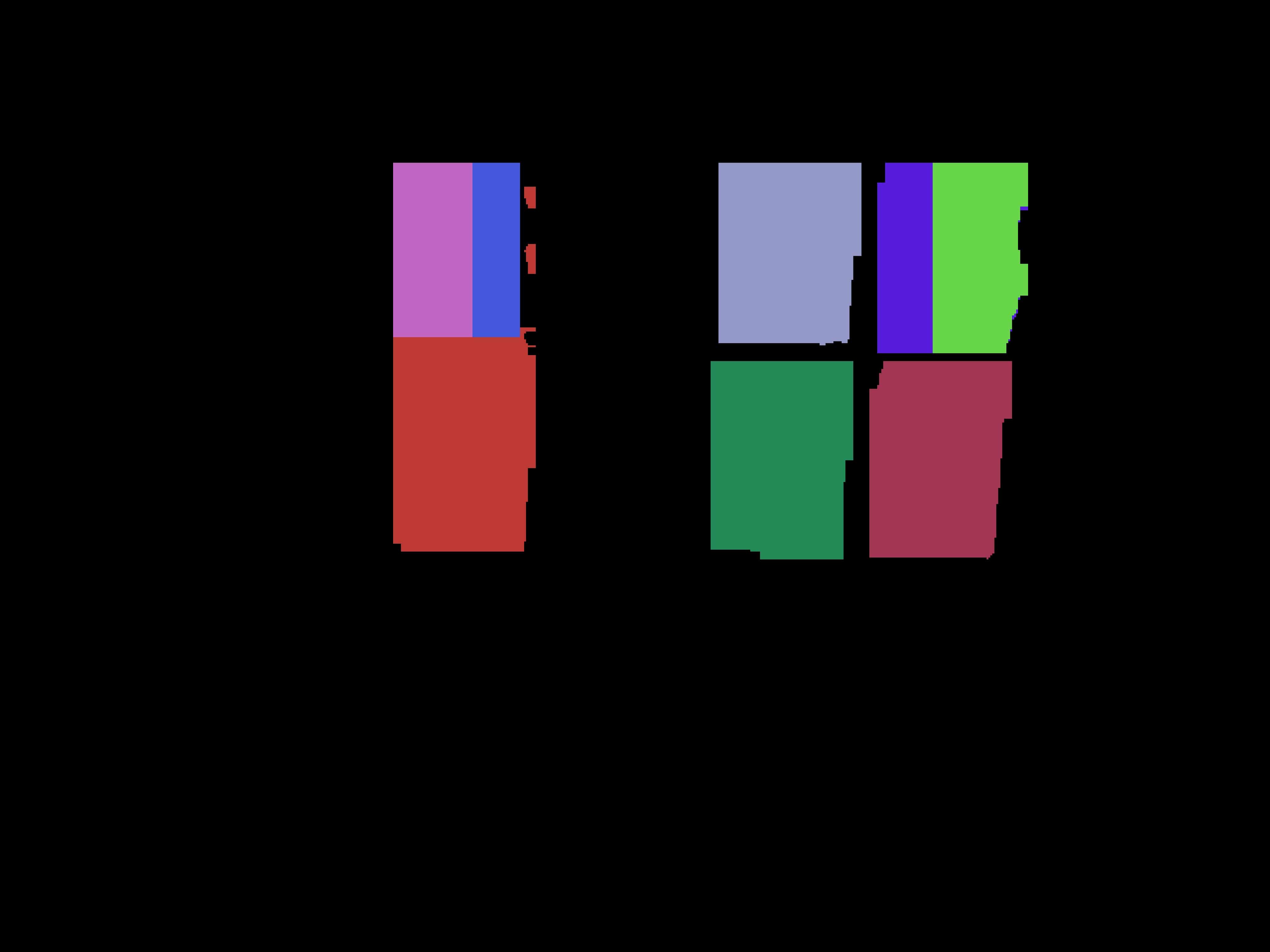} &
        \includegraphics[width=0.075\textwidth]{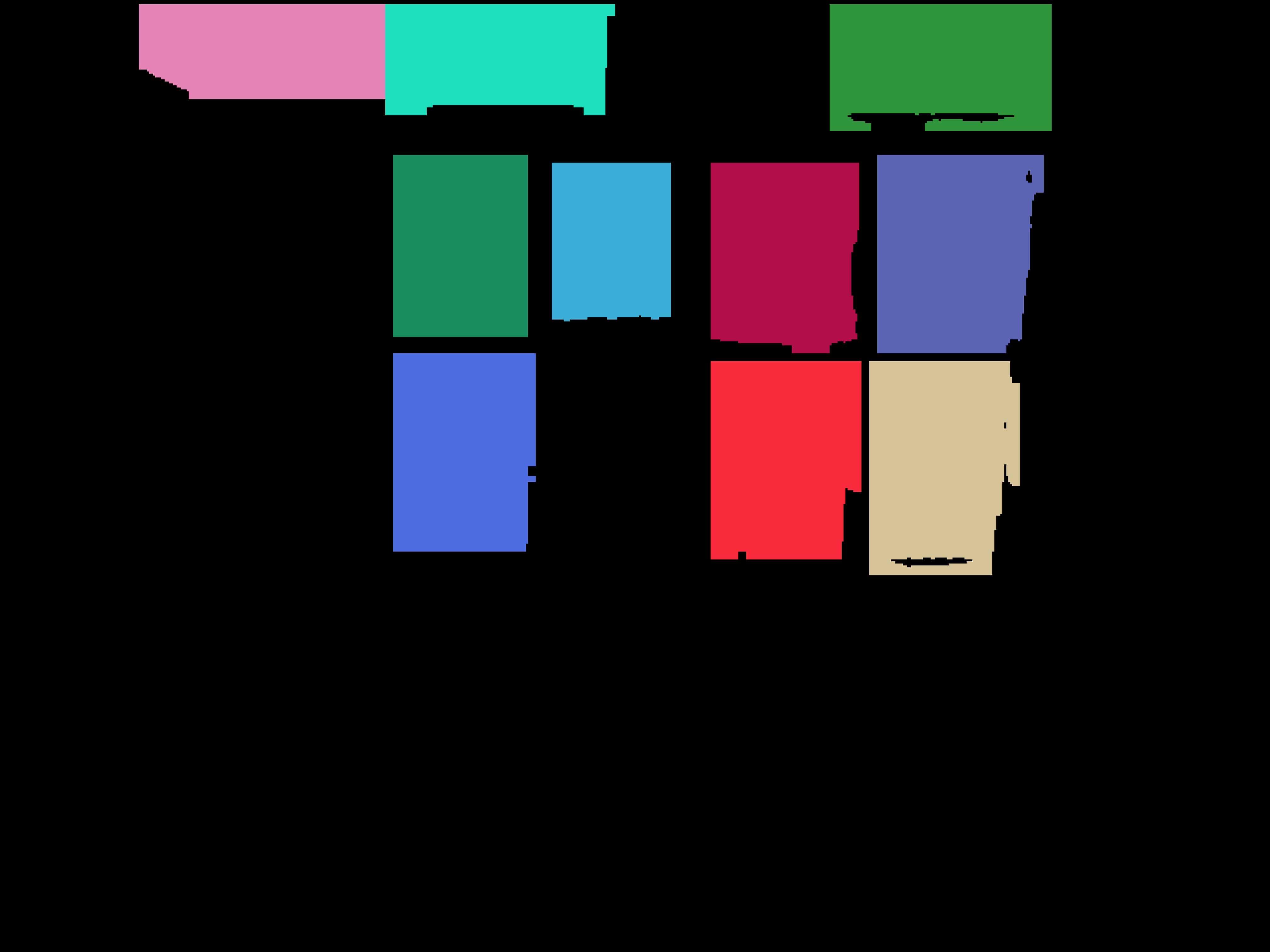} &
        \includegraphics[width=0.075\textwidth]{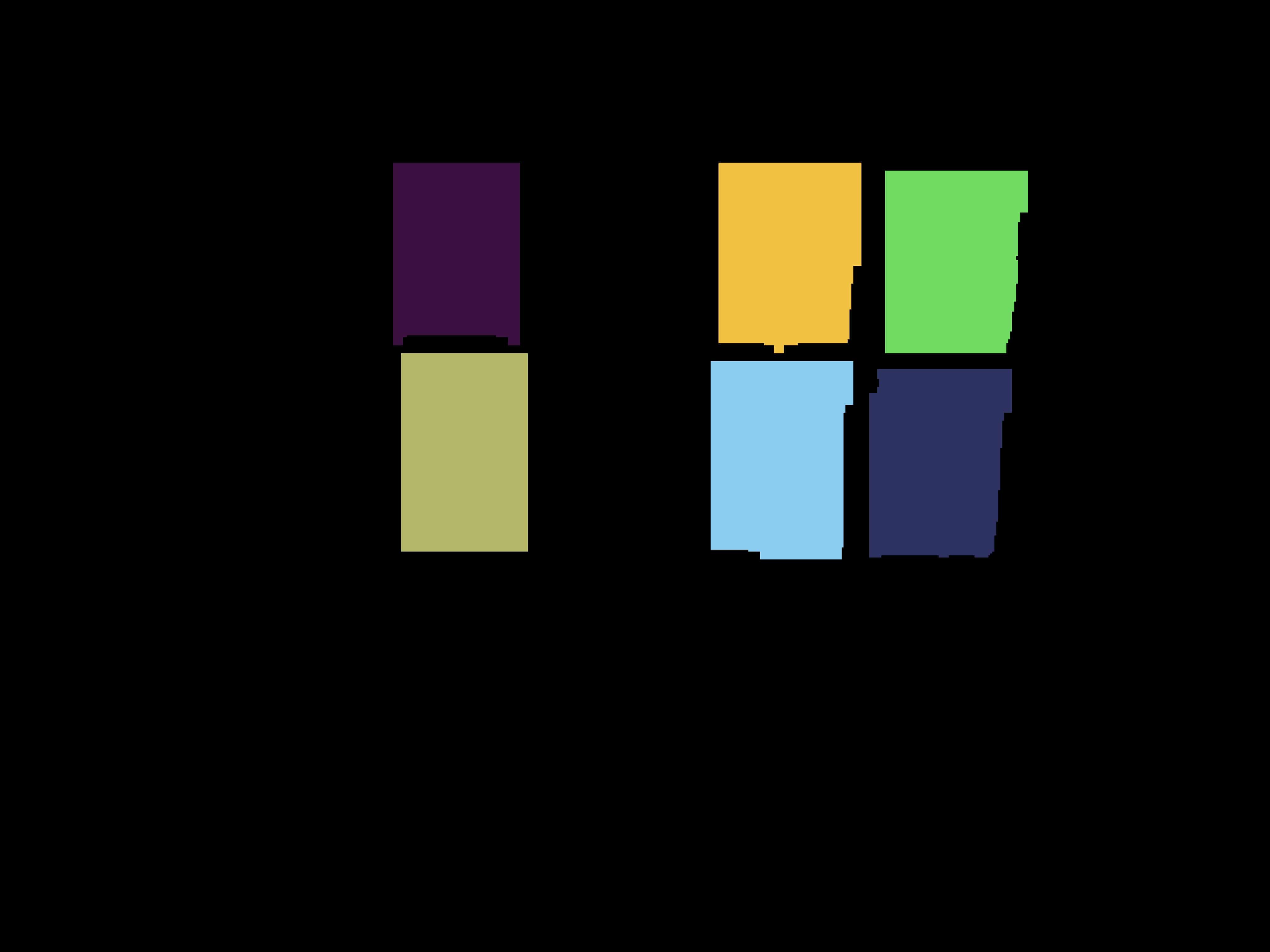} &
        \includegraphics[width=0.075\textwidth]{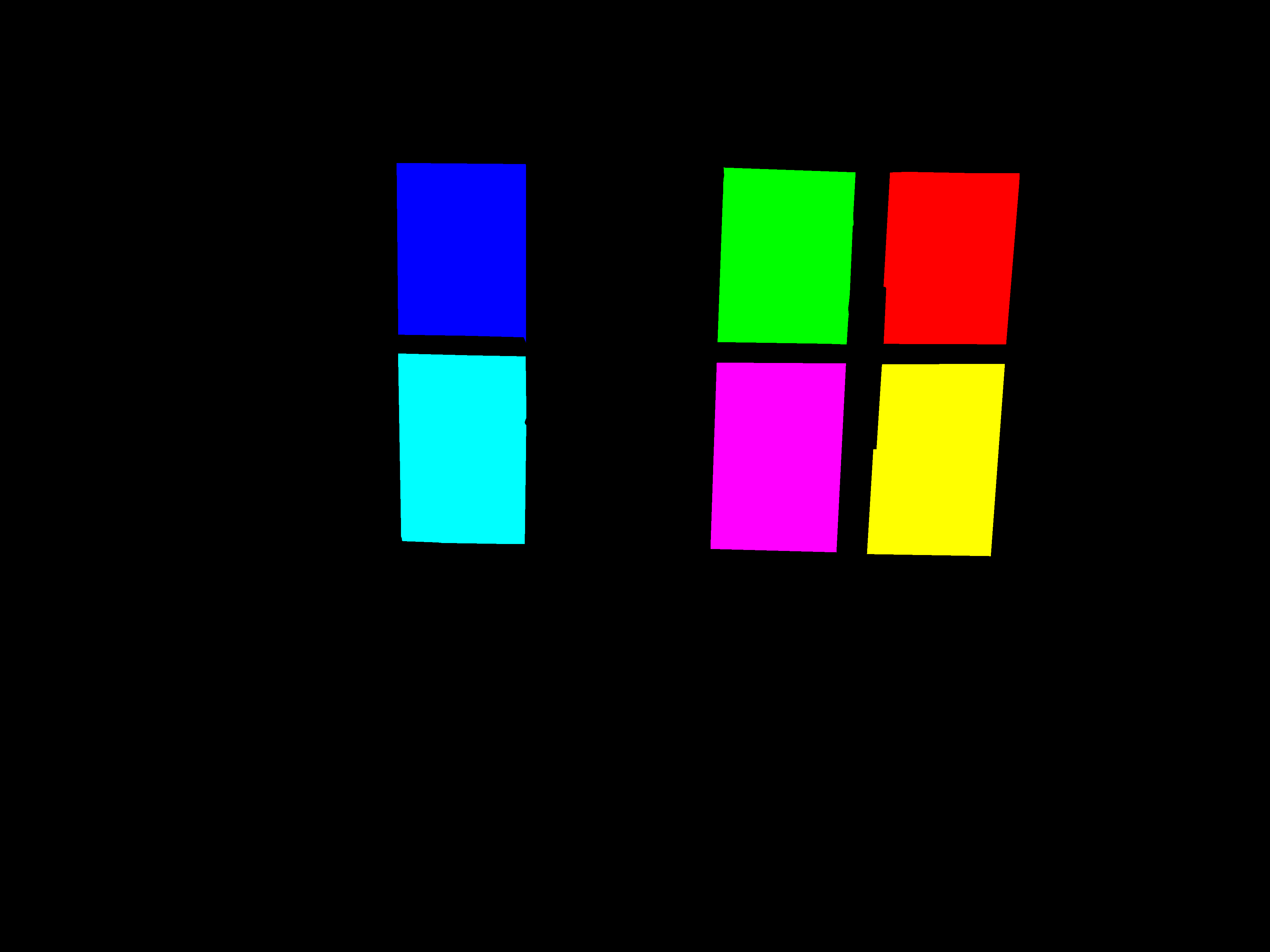} \\
        % 行间距
        
        % 第三行图片
        \includegraphics[width=0.075\textwidth]{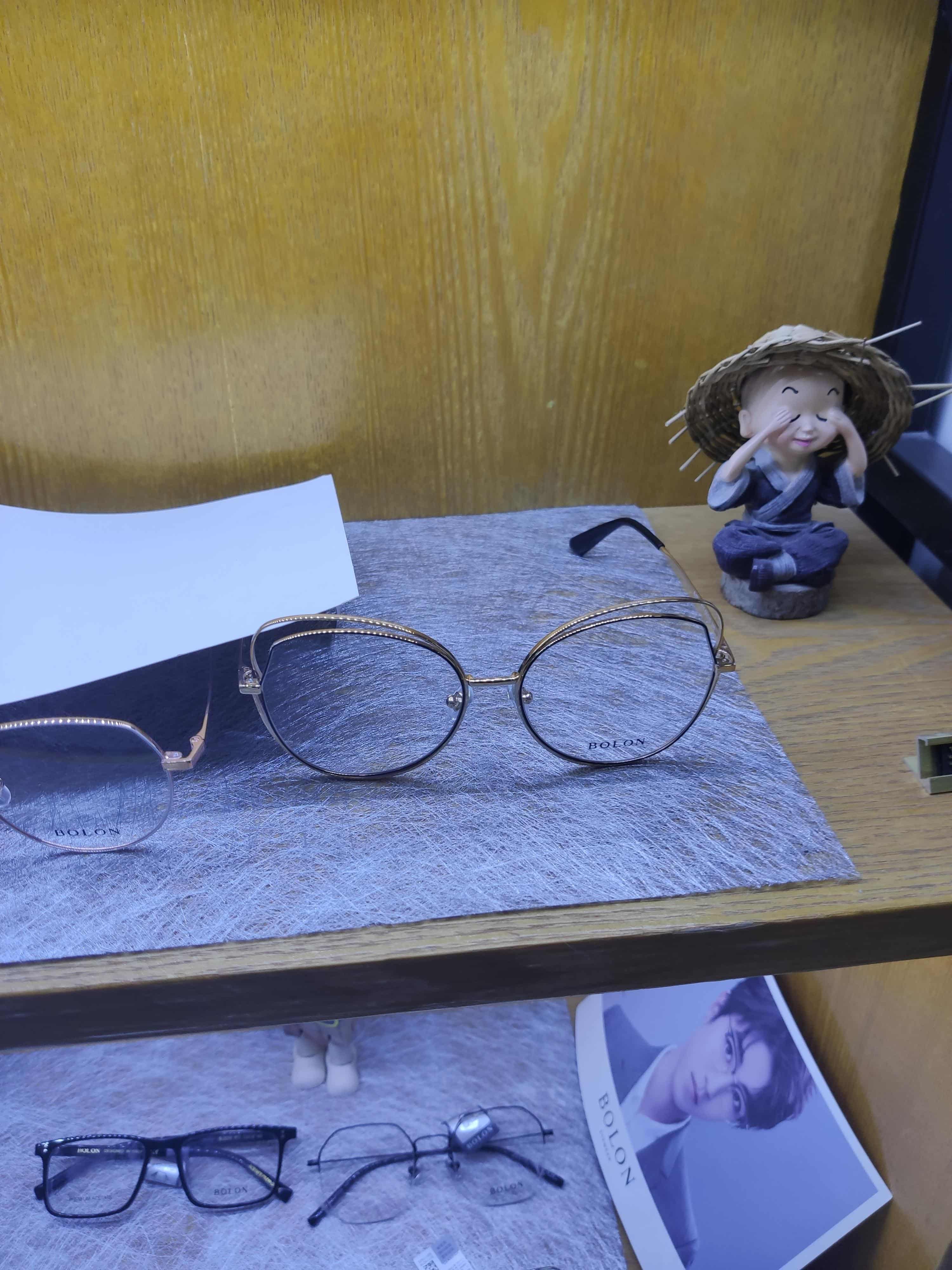} &
        \includegraphics[width=0.075\textwidth]{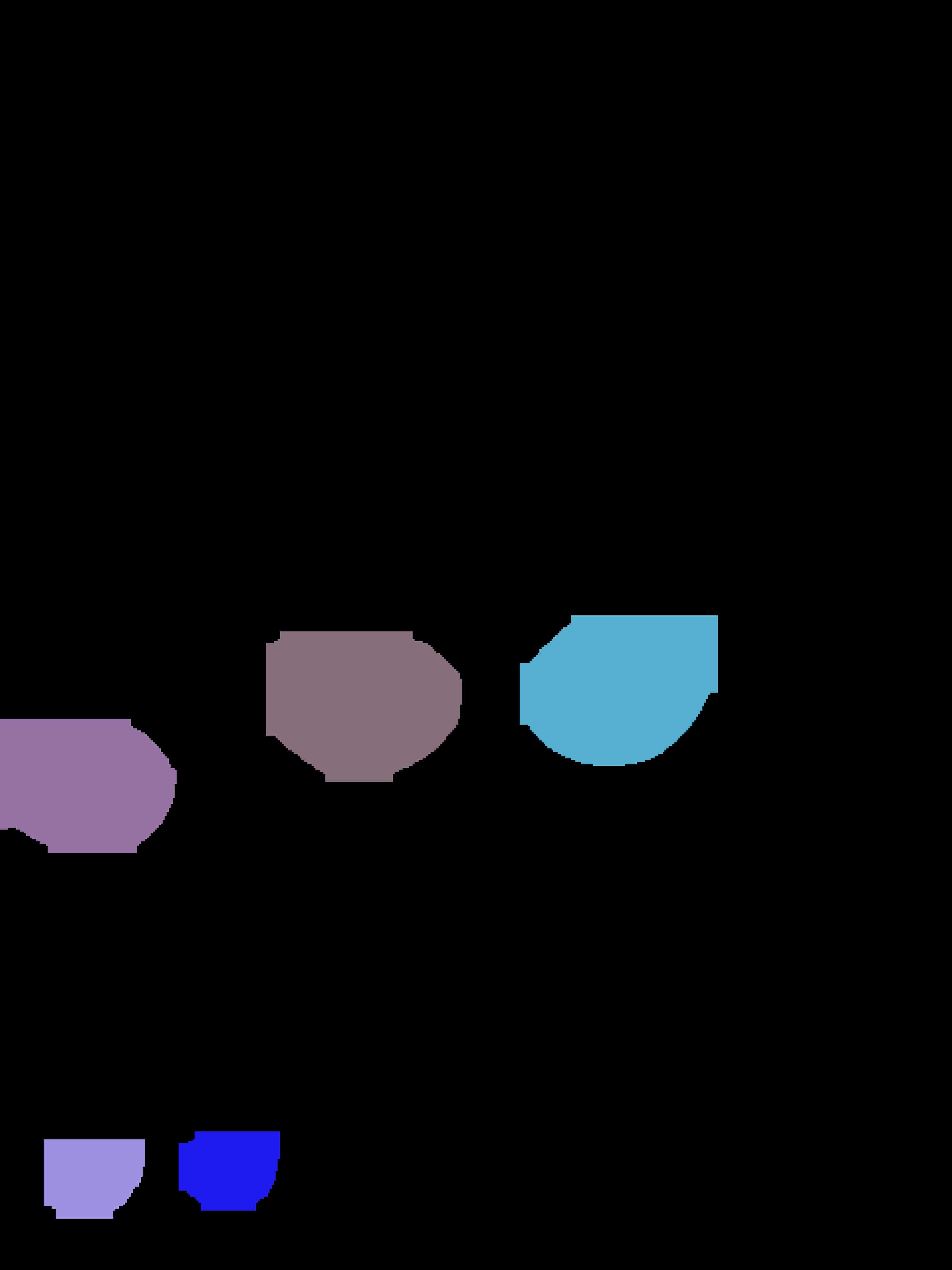} &
        \includegraphics[width=0.075\textwidth]{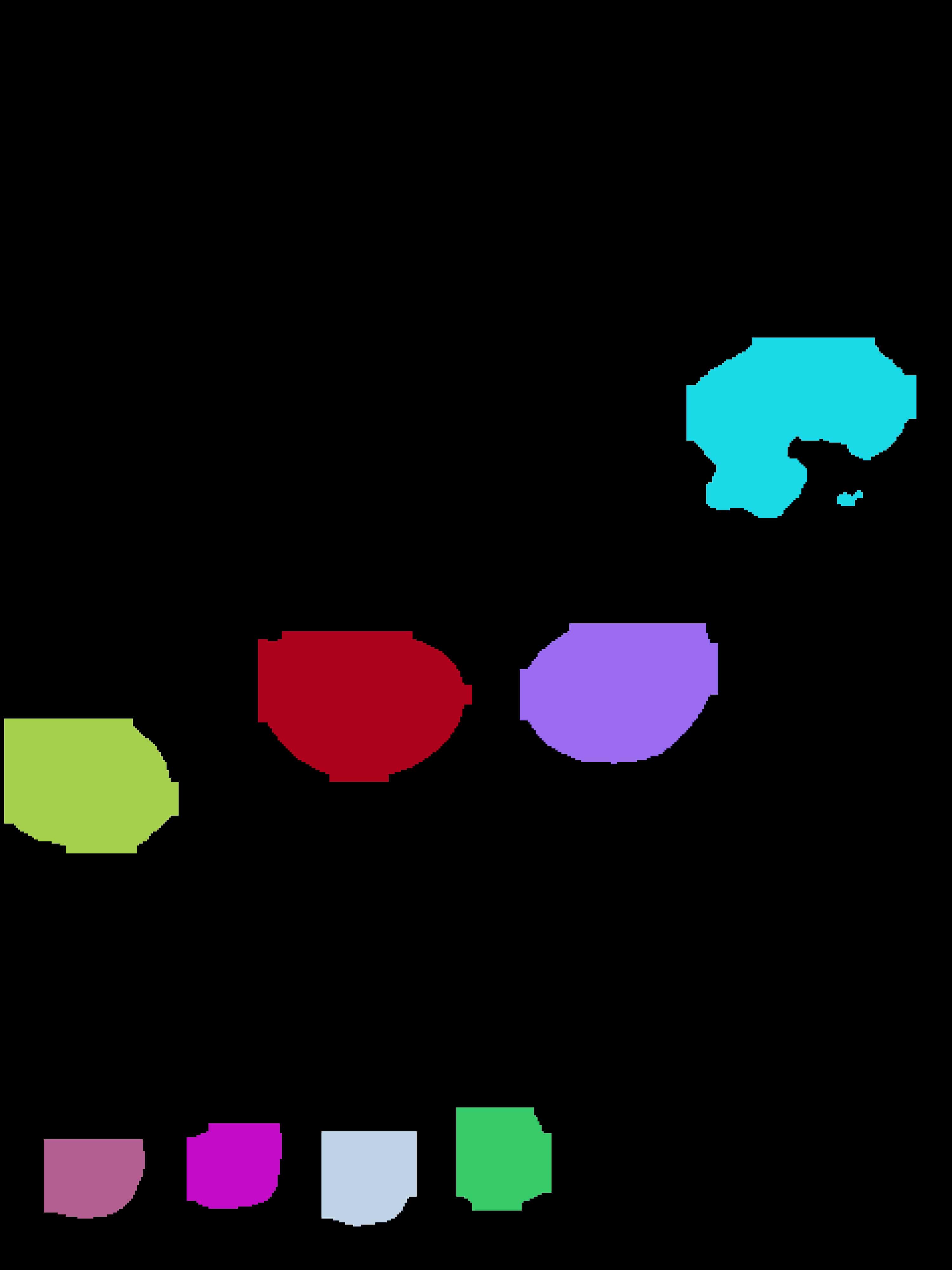} &
        \includegraphics[width=0.075\textwidth]{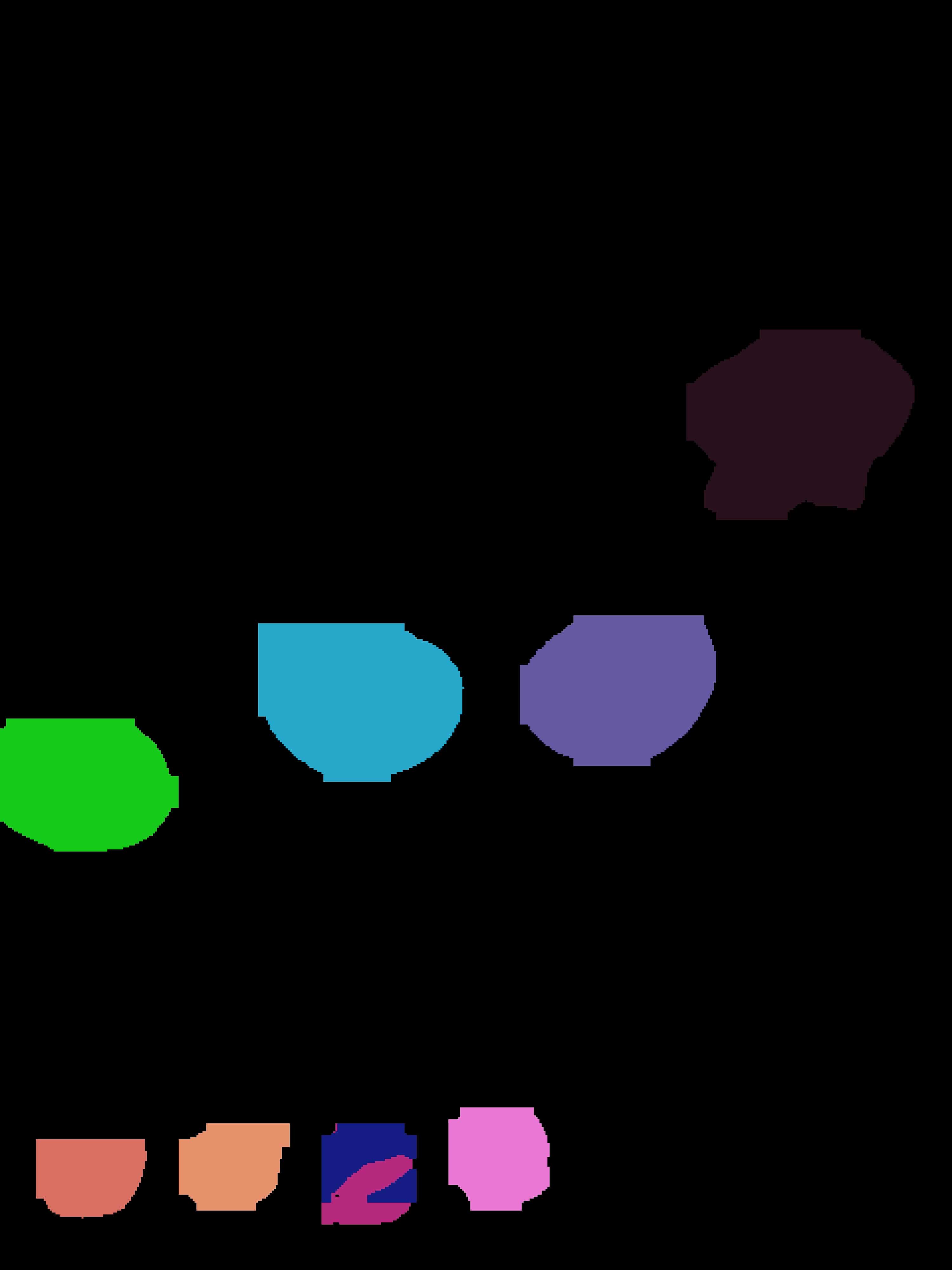} &
        \includegraphics[width=0.075\textwidth]{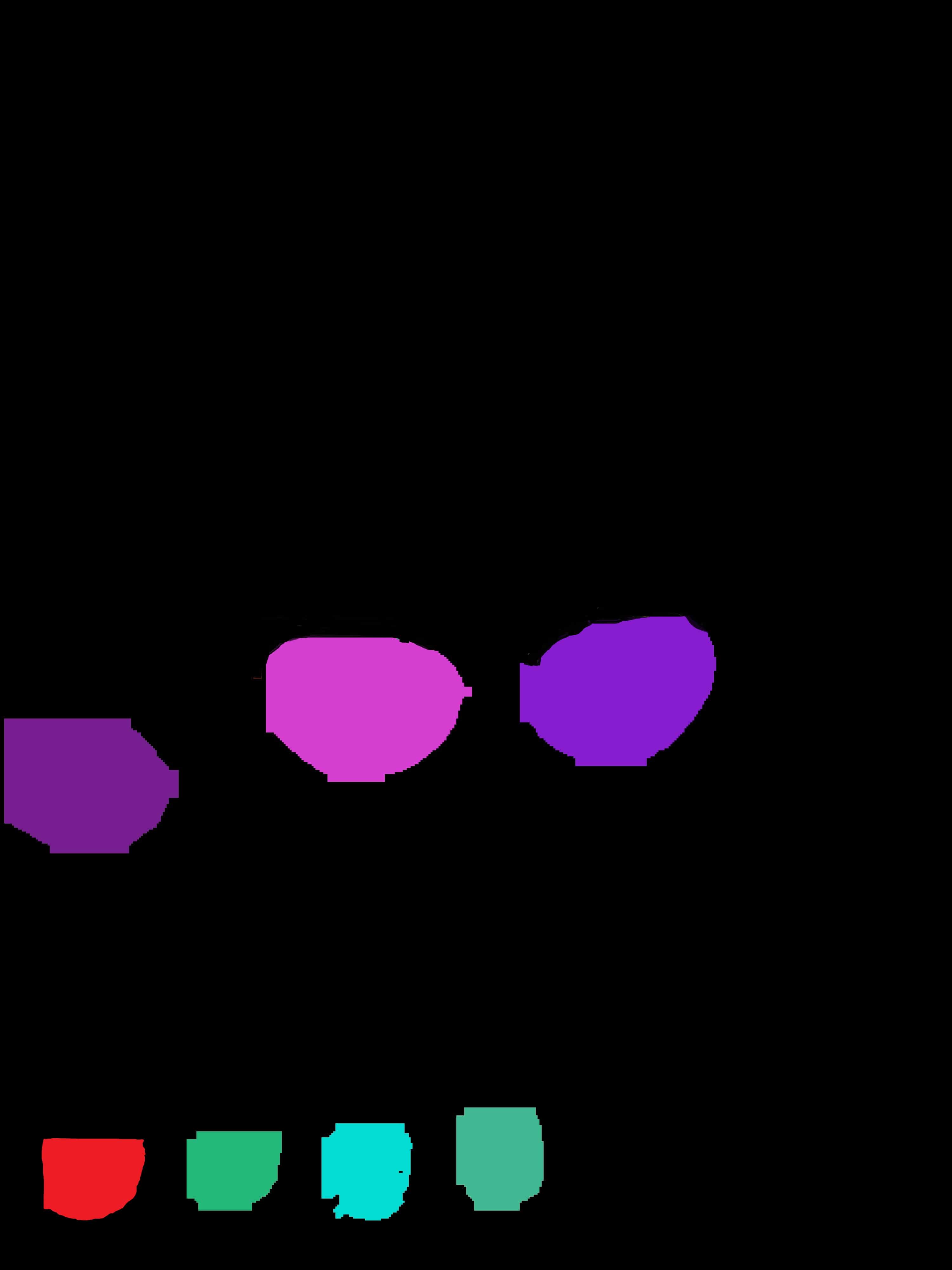} &
        \includegraphics[width=0.075\textwidth]{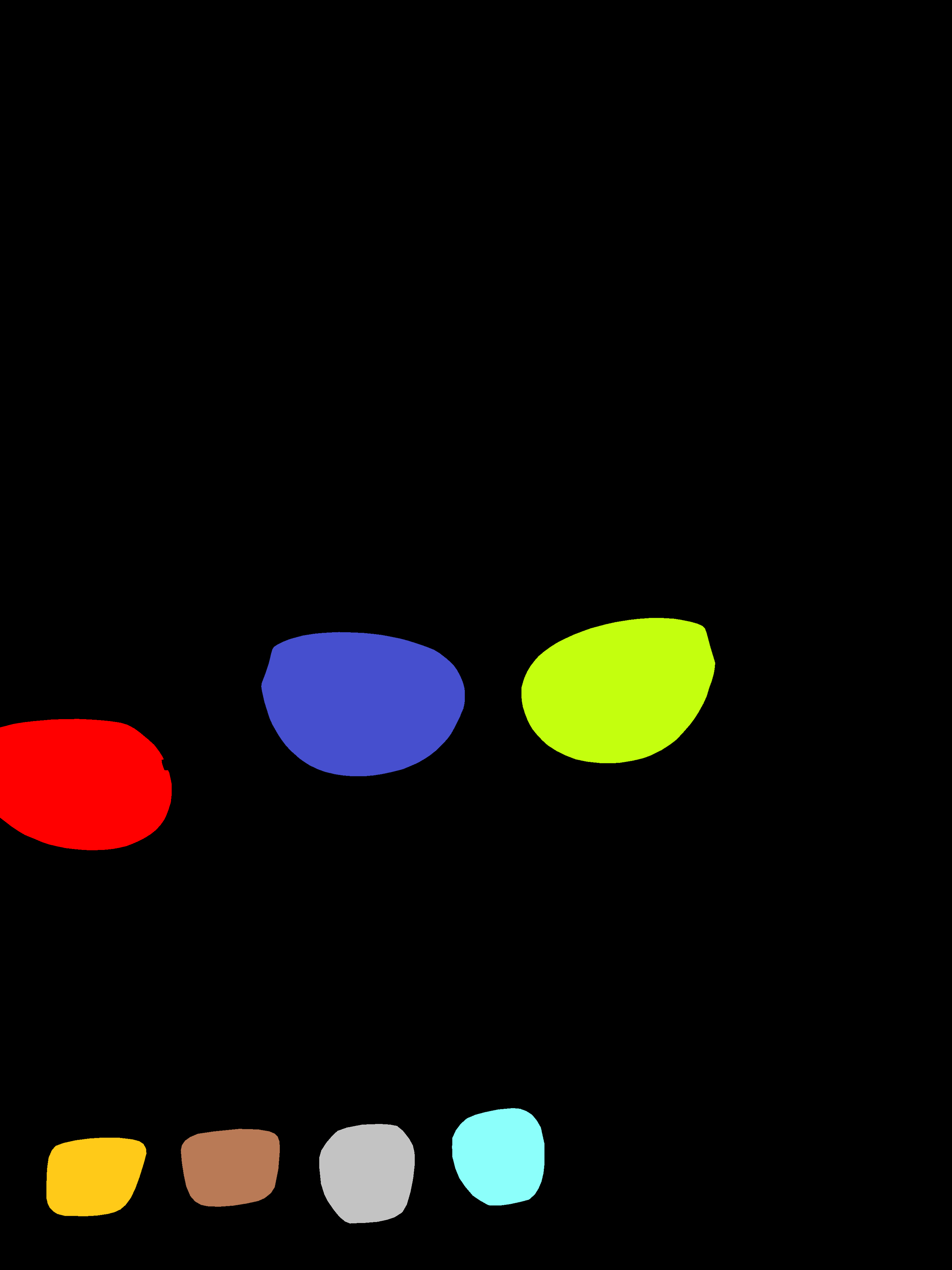} \\
        % 标签行（与图片列严格对应）
        \parbox{0.075\textwidth}{\centering Original} &
        \parbox{0.075\textwidth}{\centering yolov12} &
        \parbox{0.075\textwidth}{\centering YOLO11} &
        \parbox{0.075\textwidth}{\centering Mamba-YOLO} &
        \parbox{0.075\textwidth}{\centering Ours} &
        \parbox{0.075\textwidth}{\centering GT} \\
    \end{tabular}
    
    \caption{Visual comparison of instance segmentation results on the GVD (first row) and Trans10K (second and third rows) datasets, respectively. }
    \vspace{-1.5em}
    \label{fig:comparison_compact}
\end{figure}

These results collectively demonstrate that SEP-YOLO effectively addresses the challenges of boundary blur and low contrast in transparent object segmentation, outperforming existing methods in both accuracy and generalization. The visual comparisons in Fig.~\ref{fig:comparison_compact} further highlight its ability to produce sharper boundaries and more complete masks, especially for transparent objects with complex background interactions.

\begin{table}[htbp]
  \centering
  \caption{Ablation study of SEP-YOLO components on the Trans10K dataset. }
  \label{tab:ablation_trans10k}
  \begin{tabular}{lcccc}
    \toprule
    \multirow{2}{*}{\textbf{Method}}& \multicolumn{2}{c}{Box} & \multicolumn{2}{c}{Mask} \\
    \cmidrule(lr){2-3} \cmidrule(lr){4-5}
     & $mAP_{50}$ & $mAP_{75}$ & $mAP_{50}$ & $mAP_{75}$ \\
    \midrule
    Baseline(YOLO11) & 0.816 & 0.742 & 0.813 & 0.729 \\
    + FDDEM & 0.836 & 0.755 & 0.833 & 0.741 \\
    + FDDEM + MS-GRB & 0.845 & 0.759 & 0.841 & 0.749 \\
    + FDDEM + CA$^2$-Neck & 0.844 & 0.761& 0.840 & 0.748 \\
    SEP-YOLO (ours) & \textbf{0.852} & \textbf{0.774} & \textbf{0.851} & \textbf{0.754} \\
    \bottomrule
  \end{tabular}
  \vspace{0.2cm}
\end{table}
\begin{table}[htbp]
  \centering
  \caption{Ablation study of SEP-YOLO components on the GVD dataset. }
  \label{tab:ablation_gvd}
  \begin{tabular}{lcccc}
    \toprule
    \multirow{2}{*}{\textbf{Method}}& \multicolumn{2}{c}{Box} & \multicolumn{2}{c}{Mask} \\
    \cmidrule(lr){2-3} \cmidrule(lr){4-5}
     & $mAP_{50}$ & $mAP_{75}$ & $mAP_{50}$ & $mAP_{75}$ \\
    \midrule
    Baseline(YOLO11) & 0.851 & 0.761 & 0.849 & 0.742 \\
    + FDDEM & 0.864 & 0.773 & 0.861& 0.769 \\
    + FDDEM + MS-GRB& 0.870 & 0.784 & 0.863 & 0.758 \\
    + FDDEM + CA$^2$-Neck & 0.865 & 0.776 & 0.860 & 0.755 \\
    SEP-YOLO (ours) & \textbf{0.882} & \textbf{0.796} & \textbf{0.872} & \textbf{0.769} \\
    \bottomrule
  \end{tabular}
\end{table}
\subsection{Ablation Studies and Analysis}

Ablation studies on Trans10K and GVD datasets validate each component of SEP-YOLO. As shown in Table~\ref{tab:ablation_trans10k}, the baseline achieved 0.816 Box $mAP_{50}$ and 0.813 Mask $mAP_{50}$ on Trans10K. Adding FDDEM alone improved Box $mAP_{50}$ to 0.836 and Mask $mAP_{50}$ to 0.833, demonstrating the significance of frequency domain enhancement for transparent object boundaries. Further incorporating MS-GRB or CA$^2$-Neck provided additional gains, with the full SEP-YOLO achieving the best performance of 0.852 Box $mAP_{50}$ and 0.851 Mask $mAP_{50}$. Similar trends were observed on GVD (Table~\ref{tab:ablation_gvd}).  These results confirm the complementary role of each module in addressing the unique challenges of transparent object instance segmentation.

\section{Conclusion}
In this paper, we propose SEP-YOLO to address low contrast and boundary blurring in transparent object instance segmentation under complex scenes. Comprehensive experiments on the challenging Trans10K and GVD datasets show that SEP-YOLO significantly outperforms existing SOTA methods in segmentation accuracy, while retaining a lightweight architecture and real-time inference speed. Furthermore, we contributed high-quality instance-level annotations for the Trans10K dataset, filling a critical data gap. The robust performance and high efficiency of SEP-YOLO thus demonstrate its significant potential for industrial and robotic applications.

\newpage
\enlargethispage{\baselineskip} % 确保足够空间
\bibliographystyle{ieeetr}
\bibliography{ref}    

\end{document}